# Study of improving nano-contouring performance by employing cross-coupling controller

Wen-Yuh Jywe*, Shih-Shin Chen*, Hung-Shu Wang**, Chien-Hung Liu***
Hsin-Hung Jwo*, Yun-Feng Teng****and Tung Hsien Hsieh *****

* National Formosa University/Department of Automation Engineering, Yunlin, Taiwan R.O.C
e-mail: jywe@nfu.edu.tw
** National Cheng Kung University/Department of Mechanical Engineering, Tainan, Taiwan R.O.C
e-mail: ah.tree@gmail.com
*** National Formosa University/Institute of Electro-Optical and Materials Science, Yunlin, Taiwan R.O.C
**** National Chung-Cheng University/Department of Mechanical Engineering, Chiayi, Taiwan R.O.C
*****National Cheng Kung University/ Institute of Manufacturing Engineering, Tainan, Taiwan R.O.C

**Abstract**

For the tracking stage path planning, we design a two-axis cross-coupling control system which uses the PI controller to compensate the contour error between axes. In this paper, the stage adoptive is designed by our laboratory (Precision Machine Center of National Formosa University). The cross-coupling controller calculates the actuating signal of each axis by combining multi-axes position error. Hence, the cross-coupling controller improves the stage tracking ability and decreases the contour error. The experiments show excellent stage motion. This finding confirms that the proposed method is a powerful and efficient tool for improving stage tracking ability. Also found were the stages tracking to minimize contour error of two types circular to approximately 25nm.

## 1  Introduction

Precision positioning [1-2] and tracking motion are important developments in the field of mechanics for manufacturing products and measuring object dimensions. Basically, they can be classified into two classes of methods [2]: (a) methods to minimize contouring errors by improving axis tracking errors and (b) methods to minimize the contour error directly by using on-line calculations like in cross-coupling control systems. The latter approach, which is the topic of this paper, was originally presented by Koren [4-5], and was represented by two DDA (Digital Differential Analyzer) integrators and a digital comparator in a traditional control system. It was proved by mathematical methods that contour errors are at their minimum when the biaxial control system adopts the cross-coupling control. The variant gain cross-coupl coupling control method adjusts the gains of the controller according to the formation of its track and minimizes the contour errors under the curve path. Kulkarni [6] proposed the optimum method for controlling the track, using directly reduced contour error as a benchmark to design an optimum controller that can minimize contour errors. Srinivasan [7] discussed resonance occurring at different gains of the cross-coupl coupling controller and showed that contour errors were minimized through experiments on circle tracking. This method was used by Lue [8] to control the track of a complex 5-axle machine tool.

Careful consideration and design are required for the precision, feed-rate and loading capability of processing equipment applied to TFT LCD and optical communication so as to optimize their efficiency. Consequently, selecting optimal positioning stage technology [9-11] in processing equipment becomes one of the key factors that affect the manufacture. This research adopts the stage [12-13] as made by the Precision Machine Center of National Formosa University.

## 2  Structure and Action Principles of the Stage

The objective of this research is to control a stage that is of high precision positioning. The stage adopts a modular method to heat and then fabricate each flexure body for constructing a flexure hinge mechanism. The stage includes piezoelectric actuators (PSt1000/10/60 Vs18) and flexure bodies. The structure and object of the stage are as shown in figure 1. The measurement tools of the system are sets of capacitive sensors, and the total range of the gap sensors is $50\ \mu m$ with a sensitivity of $0.2\ V/\mu m$. The actuators are piezoelectric actuators which are of compact size, quick response, high efficiency in electro-mechanical transformation and low heat. The sensors are capacitive sensors which are contact-less, of high resolution, high response speed and high stability.

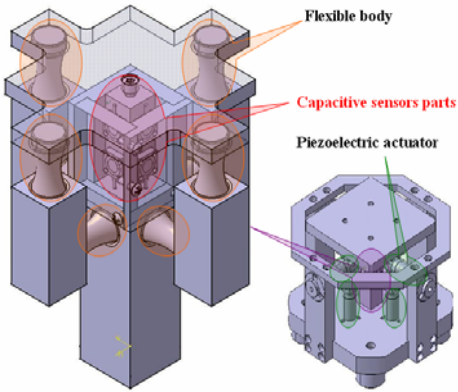

**Figure1: Structure of stage [13]**

### 3 The cross-coupling method

The objective of traditional 2-D tracking control is to reduce axial error. However, it is impossible to reduce axial error to zero due to the dynamic characteristics of the stage, which lead to contour errors under tracking behavior. This method uses the geometric relation of 2-D motion to merge axial errors of the different axes and to compensate for those errors through interdependent changes in X and Y positioning

In order to develop a cross- coupling controller, the first step is to formulate a mathematic model of contour error built up on the axial errors of X and Y that occur during motion. The mathematical model of contour error is not complicated and can be calculated quickly. Therefore, this simple cross-coupling controller is an efficient means to minimize contour error.

#### 3.1 Contour Error
In this section, controllers for contour error control are developed for circular contouring. However, the actual route differs from the command route while the stage moves, and such differences are defined as contour errors, as shown in figure 2.

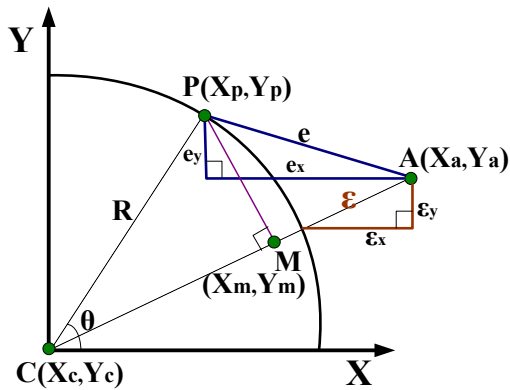

**Figure2: Contour error of curve motion in a biaxial system [13]**

Position P: the command position
Position A: the actual position of the stage's motion
Position C: the circle center
Position M: the $\overline{AC}$ and $\overline{PM}$ are perpendicular to one another point
Line R: the radius

The $e_x$ and $e_y$ are position errors along the X and Y axes respectively. $\varepsilon_x$ and $\varepsilon_y$ are contour errors that $\varepsilon$ project to the X and Y axes respectively. Angle $\theta$ is the tilt angle of the command straight line relative to the X axis, the equations for which is as follows:

$$\varepsilon = \sqrt{(X_a - X_c)^2 + (Y_a - Y_c)^2} - R \quad (1)$$

$$\begin{aligned} X_a &= X_p - e_x \\ &= X_c + R\sin\theta - e_x \end{aligned} \quad (2)$$

$$\begin{aligned} Y_a &= Y_p - e_y \\ &= Y_c + R\cos\theta - e_y \end{aligned} \quad (3)$$

Equations (2), (3) and (1) lead to:

$$\begin{aligned} \varepsilon &= \sqrt{(R\sin\theta - e_x)^2 + (R\cos\theta - e_y)^2} - R \\ &= \sqrt{R^2(\sin^2\theta + \cos^2\theta) - 2R(e_x\sin + e_y\cos\theta) + e_x^2 + e_y^2} \\ &= R\sqrt{1 - 2(\frac{e_x}{R}\sin\theta - \frac{e_y}{R}\cos\theta) + \frac{e_x^2 + e_y^2}{R^2}} - R \end{aligned} \quad (4)$$

The eq. (4) uses the binomial theorem to expansion. Because $e_x$ and $e_y$ too small, over the item $R^2$ approximate to zero (radius R is constant), can be summarized as:

$$\begin{aligned} \varepsilon &= -(e_x\sin\theta - e_y\cos\theta) + \frac{e_x^2 + e_y^2}{2R} \\ &= -(\sin\theta - \frac{e_x}{2R})e_x + (\cos\theta + \frac{e_y}{2R})e_y \end{aligned} \quad (5)$$

Then, the $e_x$ and $e_y$ excessively smaller than R, so $\frac{e_x}{2R}$ and $\frac{e_y}{2R}$ equal zero. Equation can be obtained:

$$\varepsilon = e_y\cos\theta - e_x\sin\theta \quad (6)$$

The most minimum distance of command position P and $\overline{AC}$ is point M. The length of line $\overline{AM}$ is expressed as contour error $\varepsilon$. The inference can be obtained:

Then, the slope of command path line is m, and slope $\frac{-1}{m}$ is the slope of this vertical line.

$$\frac{Y_m - Y_p}{X_m - X_p} = m \quad (7)$$
$$\Rightarrow Y_m = Y_p + mX_m - mX_p$$

$$\frac{Y_m - Y_a}{X_m - X_a} = \frac{-1}{m} \quad (8)$$
$$\Rightarrow X_m = X_a + mY_a - mY_m$$

Equations (7) and (8) can be manipulated to obtain $m(X_m, Y_m)$:

$$X_m = \frac{m^2}{1+m^2} X_p + \frac{m}{1+m^2}(Y_a - Y_p) + \frac{1}{1+m^2} X_a \quad (9)$$

$$Y_m = \frac{m^2}{1+m^2} Y_a + \frac{m}{1+m^2}(X_a - X_p) + \frac{1}{1+m^2} Y_p \quad (10)$$

The $\varepsilon_x$ and $\varepsilon_y$ can be obtained for Fig.2 by respectively describing the following:

$$\varepsilon_x = X_m - X_a$$
$$= \frac{m^2}{1+m^2}(X_p - X_a) - \frac{m}{1+m^2}(Y_a - Y_p) \quad (11)$$

$$\varepsilon_y = Y_m - Y_a$$
$$= -\frac{m^2}{1+m^2}(X_p - X_a) + \frac{1}{1+m^2}(Y_p - Y_a) \quad (12)$$

Based on the above discussion, and by the aid of equations (11) and (12), at the moment of mode switching the following matrix can be obtained by:

$$\begin{bmatrix} \varepsilon_x \\ \varepsilon_y \end{bmatrix} = \begin{bmatrix} \frac{m^2}{1+m^2} & -\frac{m}{1+m^2} \\ -\frac{m}{1+m^2} & \frac{1}{1+m^2} \end{bmatrix} \begin{bmatrix} e_x \\ e_y \end{bmatrix} \quad (13)$$

Then, the $m_x = \cos\theta$, $m_y = \sin\theta$, so equation (13) lead to:

$$\begin{bmatrix} \varepsilon_x \\ \varepsilon_y \end{bmatrix} = \begin{bmatrix} m_y^2 & -m_x m_y \\ -m_x m_y & m_x^2 \end{bmatrix} \begin{bmatrix} e_x \\ e_y \end{bmatrix} \quad (14)$$

The contour should maintain its accuracy at a certain level even when encountering disturbances. Although disturbances may cause varying parameters, the control system should not degenerate. In general, the CNC should be able to universally execute original tracking with accurate results. After proofreading your paper, it must be submitted on the ASCC2006 web site (ascc2006.tf.itb.ac.id) electronically using PDF or PS formats. Do not send hard copies or use other file formats -they will not be accepted. Proper usage of the English language is expected of all submissions (i.e., Camera-ready papers). If you submit using the PDF format, make sure that the PDF file looks fine on the screen as well as in print.

Failure to follow the above guidelines may result in a submission being rejected for publication in the conference proceedings and CD ROM.

**3.2 Structure of the Controller**
The common objective of stage motion control is to reduce axial error. However, it is impossible to reduce axial error to zero in curve motion due to a stage's dynamic characteristics. Thus, minimizing contour error during motion becomes an important objective. This research focuses on minimizing contour error by adopting a cross-coupling method to produce experiment results. figure3 shows the blocks of the cross-coupling controller when it has a closed-loop. The output signals are:

$$V_{rx} = (K_{px} + \frac{K_{Ix}}{s})e_x + K_{dx}\varepsilon_x$$
$$V_{ry} = (K_{py} + \frac{K_{Iy}}{s})e_y + K_{dy}\varepsilon_y \quad (15)$$

Axial gains $(K_{px} + \frac{K_{Ix}}{s})$ and $(K_{py} + \frac{K_{Iy}}{s})$, which are PI controllers, have the major function of improving axial errors. $K_{dx}$ and $K_{dy}$, which are the gains of the cross-coupling controller have the function of minimizing the contour errors of the stage motion.

Then, equation (17) is obtained by applying equations (15) to (16).

$$X_a = \left\{\left(K_{px} + \frac{K_{Ix}}{s} + m_y^2 K_{dx}\right)e_x - \left(m_x m_y K_{dx}\right)e_y\right\} G_x(s)$$
$$Y_a = \left\{\left(K_{py} + \frac{K_{Iy}}{s} + m_x^2 K_{dy}\right)e_y - \left(m_x m_y K_{dy}\right)e_x\right\} G_y(s) \quad (16)$$

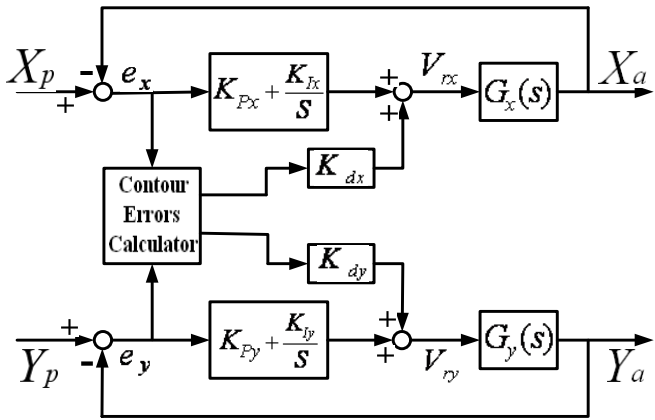

**Figure3: Blocks of the cross-coupling controller [13]**

## 4 Stage system

The experiment system shown in figure 4 includes a personal computer for the statistical algorithm and data processing, MATLAB software for compiling the control blocks, a DSP card (dSPACE company, DS1103) for executing the control program of the stage motion and processing digital/analog signals, an amplifier (model SVR 1000-3, supply of 1000 voltage) for supplying the power that the piezoelectric actuator needs to impel the stage, capacitive sensors (PI company, D-050) for sensing the axial displacement of the axes, and a signal processor for the sensors space(PI company, E-509).

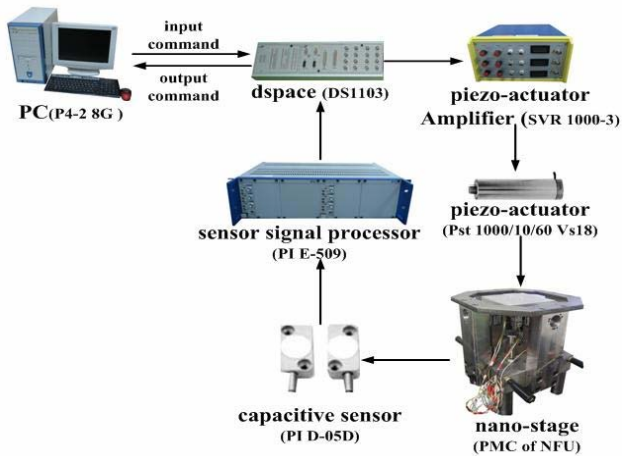

**Figure4: Flow diagram of the system [13]**

## 5 The paths design and simulation

The simulation model for the paths, and observe the influence on controller of the paths, that can do the compensation in advance to some paths in the future, or find out the best parameter of the paths to reduce the errors. The controller systems frames shown in figure 4, the blue circle mark are subsystem blocks of paths planning.

### 5.1 Tracking of curve

Simulation begins to tracking of the circular, they have two types: (1) tracking the straight line and then the circular in 45 degrees; (2) tracking the tangent line and then the circular in 45 degrees.

#### 5.1.1 Tracking the straight line and then the circular:

Simulation begins to tracking the straight line (2sec) and then the 4 circles (4sec) in 45 degrees. The input of the stage motion is frequency 1Hz, with a circle of 800nm diameter. As shown in figure 5.

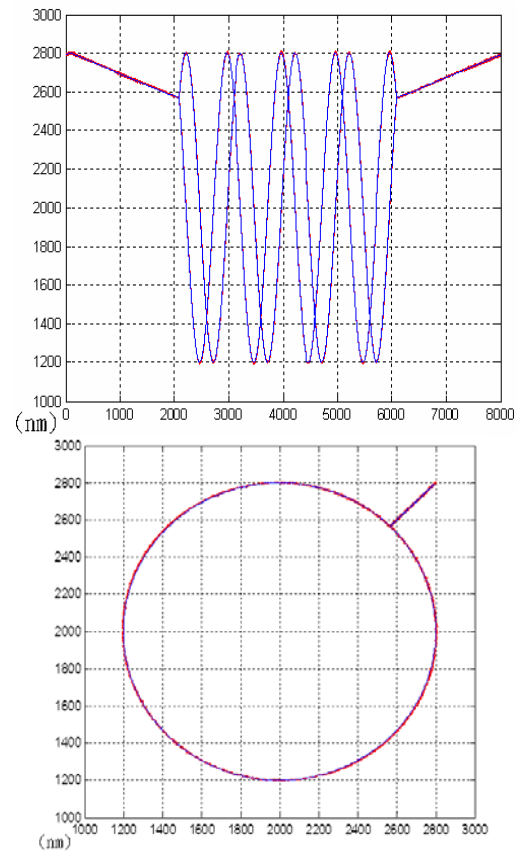

**Figure5: Tracking the straight line and then the circular[13]**

#### 5.1.2 Tracking the tangent line and then the circular:

Simulation begins to tracking the tangent line (2sec) and then the 4 circles (4sec) in 45 degrees. The input of the stage motion is frequency 1Hz, with a circle of 800nm diameter. As shown in figure 6.

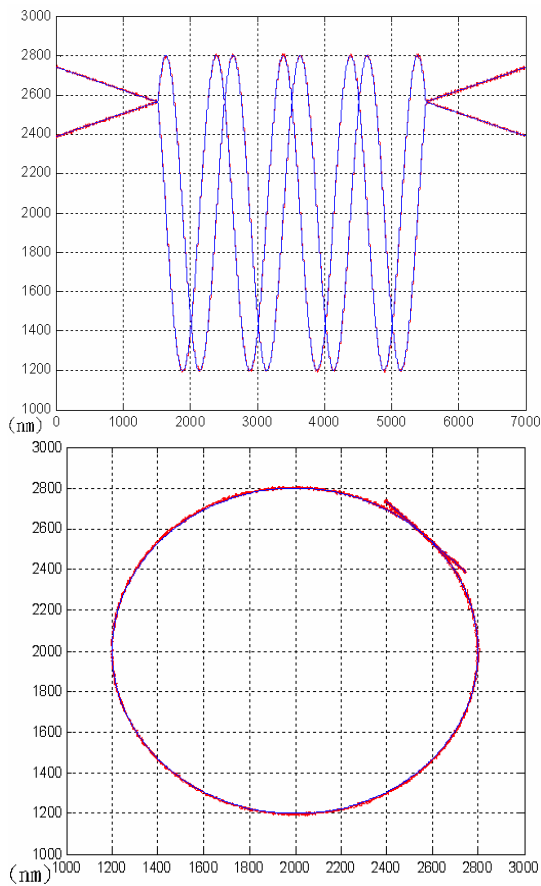

**Figure6: Tracking the tangent line and then the circular [13]**

## 6  Experiments results

The output values of the capacitor transpositioner are recorded for each trial. Errors are obtained by deducting the output value from the input value. We will discuss three types experiment in order. (1) Errors of tracking the straight line and then the circular are to approximately 25nm. As shown in figure 7. (2) Errors of tracking the tangent line and then the circular are to approximately 25nm. As shown in figure 8.

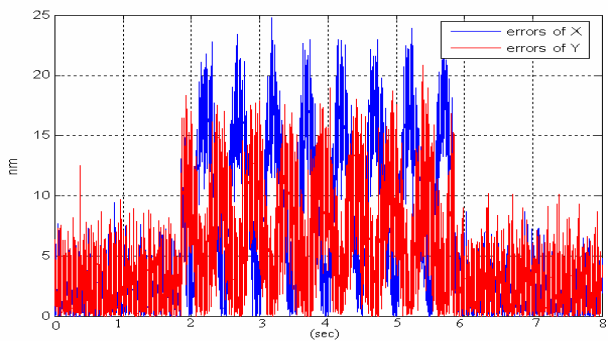

**Figure7: Errors of tracking the straight line and then the circular [13]**

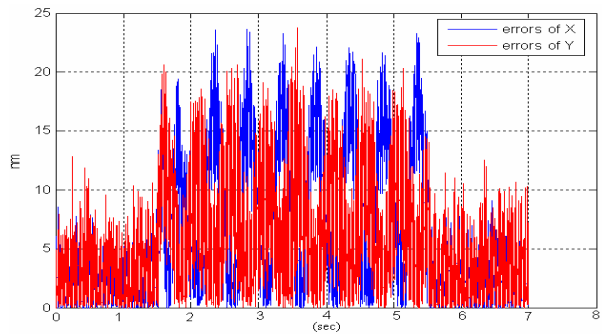

**Figure8: Errors of tracking the tangent line and then the circular [13]**

## 7  Conclusions

The precision positioning and tracking motion of a stage are important developments in the field of mechanics. The experimental results show that the stage reduces remarkably the contour error by adopting cross-coupling control theory. The conclusions are as follows:

(1) Combine a PI controller with a cross-coupling controller to improve the precision positioning and tracking motion of a stage.

(2) The simulation is to utilize simple MATLAB-SIMULINK block not to need the difficult program.

(3) The simulation model for the paths, which can do the compensation in advance to some paths in the future, or find out the best parameter of the paths to reduce the errors.

(4) The experiments of two types described in this research minimized contour error to approximately 25nm in frequency 1HZ, radius 800nm.

**Acknowledgment**

The work was supported by the National Science Council, Taiwan, Republic of China (94-2212-E-150-013 & NSC 94-2212-E-150-007).

### References

[1] Perng, M. H., and Wu, S. H., "A fast control law for nano-positioning", International Journal of Machine Tools & Manufacture, Vol. 46, pp. 1753 – 1763, 2006.

[2] Stephane Ronchi, Olivier Company, Sebastien Krut, Francois Pierrot and Alain Foumier, "High Resolution Flexible 3-RRR Planar Parallel Micro-Stage in Near Singular Conf1guration for Resolution Improvement", ICIT(IEEE International Conference), pp. 395-400, 2005.

[3] Chin, J. H., Cheng, Y. M., and Lin, J.H., "Improving contour accuracy by fuzzy-logic enhanced cross-

coupled pre-compensation method", Robotics Comp.-Integrated Mf., Vol. 20, pp. 65−76, 2004.

[4] Koren, Y., "Cross-coupled biaxial computer control for manufacturing", ASME Transactions, Journal of Dynamic Systems, Measurement and Control Vol.112, pp. 225 -232, 1980.

[5] Koren Y. and Lo Ch.-Ch., "Variable-Gain Cross-Coupling Controller for Contouring", Annals of the CRIP, Vol. 40, pp. 371-374, 1991.

[6] Kulkarni P. K. and Srinivasan K., "Cross-Coupled Control of Biaxial Feed Drive Servomechanisms", Journal of Dynamic Systems, Measurement, and Control-Transactions of the ASME, Vol. 112, pp. 225-232, 1990.

[7] Srinivasan K. and Tsao T. C., "Machine Tool Feed Drives and their Control-A Survey of the State of the Art", Journal of Manufacturing Science and Engineering, Vol. 119, pp. 743-748, 1997.

[8] Lue C. W., "A Study on the Coordinated motion of CNC machine and motion simulation platform", National Chiao-Tung University, Taiwan, ROC. Master's Thesis ,2002.

[9] Liu Chien Hung, and Jywe Wen-yuh, "A four-degree-of-freedom micro-stage for the compensation of eccentricity of a roundness Measurement machine", International Journal of Machine Tools & Manufacture, Vol. 44, pp. 365-371, 2004

[10] Oh Jeong Seok, Bae Eun Deok, Keem Taeho, and Kim Seung-Woo, "Measuring and compensating for 5-DOF parasitic motion errors in translation stages using Twyman−Green interferometry", International Journal of Machine Tools & Manufacture, Vol. 46, pp. 1748-1752, 2006

[11] Liguo Chen, Weibin Rong, Lining Sun, and Hui Xie, "Micromanipulation Robot for Automatic Fiber Alignment", International Conference on Mechatronics & Automation (Proceedings of the IEEE), Vol. 4, pp. 1756 - 1759, 2005.

[12] Jywe W. Y., Shen J. C., Liu C. H., Deng Y. F., and Jian Y. T., "Development of a Novel Heavy-Loading Flexure Hinge Based Stack-Type Five-Degree-of-Freedom Nanometer Positioning Stage", Journal of Chinese Society of Mechanical Engineering, Vol. 25, pp. 465~474, 2004

[13] Wang Hung Shu, "Optimization of cross-coupled controller parameters for nano-stage tracking motion by using the Taguchi method", National Formosa University, Taiwan, ROC. Master's Thesis, 2006.